\documentclass[conference]{IEEEtran}
\IEEEoverridecommandlockouts
\usepackage{graphics} % for pdf, bitmapped graphics files
\usepackage{graphicx} % for pdf, bitmapped graphics files
\usepackage{mathptmx} % assumes new font selection scheme installed
\usepackage{times} % assumes new font selection scheme installed
\usepackage{amsmath} % assumes amsmath package installed
\usepackage{amssymb}  % assumes amsmath package installed
\usepackage{color}
\usepackage{subfig}
\usepackage{epstopdf}
\usepackage{balance}
\usepackage{wrapfig}
\usepackage{adjustbox}
\usepackage{booktabs, makecell}
\usepackage{multirow}
\usepackage{times}
\usepackage{textcomp}
\usepackage{caption}
\DeclareMathAlphabet{\mathcal}{OMS}{cmsy}{m}{n}
\usepackage[normalem]{ulem}
\usepackage{hhline}
\usepackage{diagbox}
\usepackage{balance}
\usepackage{array}
\usepackage{float}
\usepackage{wrapfig,lipsum,booktabs}
\usepackage{graphicx} % remove [demo] in your file

\newcolumntype{L}{>{\centering\arraybackslash}m{1.3cm}}

\def\BibTeX{{\rm B\kern-.05em{\sc i\kern-.025em b}\kern-.08em
    T\kern-.1667em\lower.7ex\hbox{E}\kern-.125emX}}

\begin{document}

\title{Hardware Acceleration of Monte-Carlo Sampling for Energy Efficient Robust Robot Manipulation
\thanks{This work is supported by equipment grants from Nvidia and Xilinx Corporations, and a grant through the Brown University Office of Research Development.}
}

\author{\IEEEauthorblockN{Yanqi Liu}
\IEEEauthorblockA{\textit{Dept.\/of Computer Science} \\
\textit{Brown University}\\
Providence, RI, USA \\
yanqi\_liu@brown.edu}
\and
\IEEEauthorblockN{Giuseppe Calderoni}
\IEEEauthorblockA{\textit{Dept.\/of Automation and Informatics} \\
\textit{Politecnico di Torino}\\
Torino, Italy \\
giuseppe.calderoni@studenti.polito.it}
\and
\IEEEauthorblockN{R. Iris Bahar}
\IEEEauthorblockA{\textit{School of Engineering} \\
\textit{Dept. of Computer Science} \\
\textit{Brown University}\\
Providence, RI  USA \\
iris\_bahar@brown.edu}
}

\maketitle

\begin{abstract}
Algorithms based on Monte-Carlo sampling have been widely adapted in robotics and other areas of engineering due to their performance robustness. However, these sampling-based approaches have high computational requirements, making them unsuitable for real-time applications with tight energy constraints. In this paper, we investigate 6 degree-of-freedom (6DoF) pose estimation for robot manipulation using this method, which uses rendering combined with sequential Monte-Carlo sampling. While potentially very accurate, the significant computational complexity of the algorithm makes it less attractive for mobile robots, where runtime and energy consumption are tightly constrained.
%inefficient to implement on an embedded platform for mobile robot manipulation.
To address these challenges, we develop a novel hardware implementation of Monte-Carlo sampling on an FPGA with lower computational complexity and memory usage, while achieving high parallelism and modularization. Our results show 12X--21X improvements in energy efficiency over low-power and high-end GPU implementations, respectively. Moreover, we achieve real time performance without compromising accuracy.

\end{abstract}


\section{Introduction}
\label{sec:intro}

Robot manipulation tasks generally involve three stages: object recognition, pose estimation, and object manipulation. Convolutional Neural Networks (CNNs) have shown high accuracy and fast inference speed in object recognition, making their use widely popular for robotic applications, such as picking up and manipulating objects. However, CNNs also have several shortcomings, including extensive training effort, opacity in decision making and inability to recover from incorrect decisions. Moreover, CNNs tend to overfit to the training data due to their high non-linearity and parameter counts~\cite{srivastava2014dropout}. Overfitting also makes the CNN vulnerable to adversarial attack (e.g., via small image perturbations~\cite{goodfellow2014explaining},~\cite{eykholt2018robust}), and can also lead to poor predictions when faced with unfamiliar scenarios. In particular, when the robot operates in the real world, it is subject to complex and changing environments that often have not been captured by training data.  

Alternatively, discriminative-generative algorithms~\cite{sui2017sum},~\cite{narayanan2016discriminatively}, \cite{grip} offer a promising solution to achieve robust performance. Such methods combine the discriminative power of inference (using deep neural networks) with generative Monte-Carlo sampling to achieve robust and adaptive perception. In particular, the Monte-Carlo sampling stage can recover from the false negatives obtained from neural network outputs and offers an explainable final decision. For instance, the discriminative-generative approach of~\cite{grip} demonstrated over a 50\% improvement in pose estimation accuracy compared to end-to-end neural network approaches, which enables robust robot manipulation under various environmental changes. However, while neural network inference can be completed within a second on modern general purpose graphic processing units (GPUs), the iterative process of Monte-Carlo sampling does not map well to GPU acceleration, making the algorithm less amenable to meeting the energy and real-time constraints required of mobile applications. In particular, the run time and energy consumption is determined by the range of sampling, the number of iterations, and the computational complexity of the likelihood function.
Instead, some other means of hardware acceleration is required to make Monte-Carlo sampling fast as well as energy efficient.

%In order to improve the runtime of the discriminative-generative algorithm, general purpose graphic processing units (GPUs) have been utilized, as they operate at high clock frequencies and can achieve high computational parallelism.
%GPUs consume significant energy, and are not specifically optimized for the required computation. 
Custom hardware implementations (using FPGAs or ASICs) can operate with reduced energy consumption, even while running at a lower clock frequency, since they have better dataflow flexibility than GPUs or CPUs. However, a direct translation from the software implementation to hardware often is hardly able to yield any improvements. This paper describes a novel FPGA implementation of Monte-Carlo sampling that provides the same accuracy as GPU-CPU approaches such as~\cite{sui2018never}, but with significantly improved runtime and energy consumption.
%can be employed for object pose estimation for robot manipulation. 
This paper makes the following contributions:
\begin{itemize}
    \item We develop a complete Monte-Carlo generative inference flow suitable for hardware acceleration on an FPGA.
    \item We demonstrate how pipelining, numerical quantization, partial rasterization, and image storage optimizations can be used to significantly reduce computational complexity and memory utilization of the generative algorithm. 
    \item We show how to partition the algorithm using multiple parallel customized processing cores to increase throughput and memory access efficiency.
    \item We show that our FPGA Monte-Carlo sampling design achieves a 12X--21X improvement in energy efficiency compared to GPU-CPU implementations, while providing real time performance with no accuracy loss. 
    %\zsui{There can be one more contribution here: the much less power consumption over the comparable pose accuracy }
\end{itemize}

\section{Background}
%\subsection{Robot Manipulation}

Robot perception is an important step for robot manipulation in unstructured environments. In particular, object pose estimation is the key step for robot manipulation. Learning-based methods have been used based on end-to-end neural networks. For instance, PoseCNN~\cite{posecnn} proposes an network that learns the object segmentation with 3D translation and rotation. DOPE~\cite{dope} focuses on performance in dark environments by training on synthetic data from domain randomization and photo-realistic simulation. DenseFusion~\cite{densefusion} concatenates features extracted from object segmentation and point clouds to estimate the pose from this hybrid RGB-depth representation of the object. While the end-to-end network based methods can achieve real time performance on GPUs, they require relatively large training sets for 6 DoF object poses. Moreover, the network accuracy may be severely affected under challenging natural environments (e.g. change of lighting conditions and objects occlusion) as evaluated in~\cite{grip}.

In this paper, we focus on discriminative-generative methods, where neural network output is followed with a probabilistic inference in a two-stage paradigm. This approach is fundamentally different from the end-to-end learning network approaches proposed in~\cite{posecnn,dope,densefusion}, where the performance largely depends on network accuracy, so there is no way to recover once it has made a false decision. Techniques based on discriminative-generative methods include the work of~\cite{sui2018never}, which proposed to use a pyramidCNN to generate a probability heatmap of the object, followed by a bootstrap filter to find the optimal object pose from the object distribution. GRIP~\cite{grip} further improves the performance of~\cite{sui2018never} in dark, occluded scenes by exploiting point cloud features.

%\vspace*{2mm}
%\subsection{FPGA Acceleration}
%Pose estimation has been implemented on an FPGA in~\cite{sensorfusion} using inputs from IMU and GPS sensor measurements, and a Kalman filter to generate sensor fusion of visual odometry. The IMU and GPS sensor data is at a much lower complexity than RGB image data in our application. Also, the Kalman filter is a much lower computational complexity compared to Monte Carlo sampling \zsui{So is this better to use IMU, GPS and Kalman Filter to do on-board pose estimation? }. 
%The work of~\cite{fpga-icp} implements an FPGA-accelerated iterative closest point (ICP) algorithm to estimate the pose of the object. Iterative closest point computes the closest point distance between the source model with the target model and applies the transformation generated by the matching point to source model and continues this process iteratively to minimize the point distance error. \cite{fpga-icp} implements a sorting network for parallel point-distance comparison. 
%Our method employs a same idea; however, we take a generative approach to represent our pose assumption and therefore, our algorithm is more robust but the at the same time, more computationally expensive. In general, we found few works that consider accelerating pose estimation for robot manipulation and believe accelerating this step is very important for fast runtime and power efficiency.

Pose estimation is an important step for real-time systems, yet there is little work that considers how it may be accelerated in hardware, and these approaches are either not accurate enough for such tasks as robot manipulation~\cite{sensorfusion}, provide only partial solutions (e.g.,~\cite{schaeferling2012object},~\cite{konomura2016fpga}), or cannot be integrated with a discriminative-generate approach~\cite{fpga-icp}, which is especially useful for reasoning in unstructured environments.

A generative Monte-Carlo approach provides a greater search space and explainable reasoning, which improves accuracy and robustness, but at the expense of computational complexity. 
Particle filtering (an application of Monte-Carlo sampling) has been implemented on FPGAs for accelerating object tracking and robot mapping and localization~\cite{murai2019visual},~\cite{10.1007/978-3-319-16214-0_8},~\cite{7560264}, though not for pose estimation. Our goal is to develop a novel FPGA design for 6 DoF object pose estimation based on Monte-Carlo sampling that achieves real time performance with significantly reduced energy consumption.

%Accelerating pose estimation is a very important step for real-time robotics.  Yet, there are few works that consider its acceleration for robot manipulation. Four approaches that we are aware of are described in~\cite{sensorfusion}, \cite{schaeferling2012object}, \cite{konomura2016fpga} and~\cite{fpga-icp}. In particular, pose estimation has been implemented on an FPGA in~\cite{sensorfusion} using inputs from IMU and GPS sensor measurements, and a Kalman filter to generate visual odometry of the object. 
%The method does not consider object class and performs with less precision than required for robot manipulation.  The works of~\cite{schaeferling2012object} and~\cite{konomura2016fpga} both use the idea of image feature extraction.  In particular,~\cite{schaeferling2012object} performs SURF-based feature extraction and matching of a stereo camera on an FPGA and uses a CPU to perform 3D feature reconstruction and a RANSAC algorithm to estimate the pose of the camera. Konomura \textit{et al}.~\cite{konomura2016fpga} measure a quadcoptor pose utilizing non co-planar markers on the ground and uses an FPGA to extract the outline of the markers followed by a CPU to process the marker vertices' location and generate the pose of the quadcoptor. Both methods only use the FPGA for feature extraction but the actual pose estimation is done on the CPU. An FPGA-accelerated iterative closest point (ICP) algorithm to estimate the object pose is implemented in~\cite{fpga-icp} using a sorting network for parallel point-distance comparison.

\begin{figure}[t]
    \centering

    \includegraphics[width=\columnwidth]{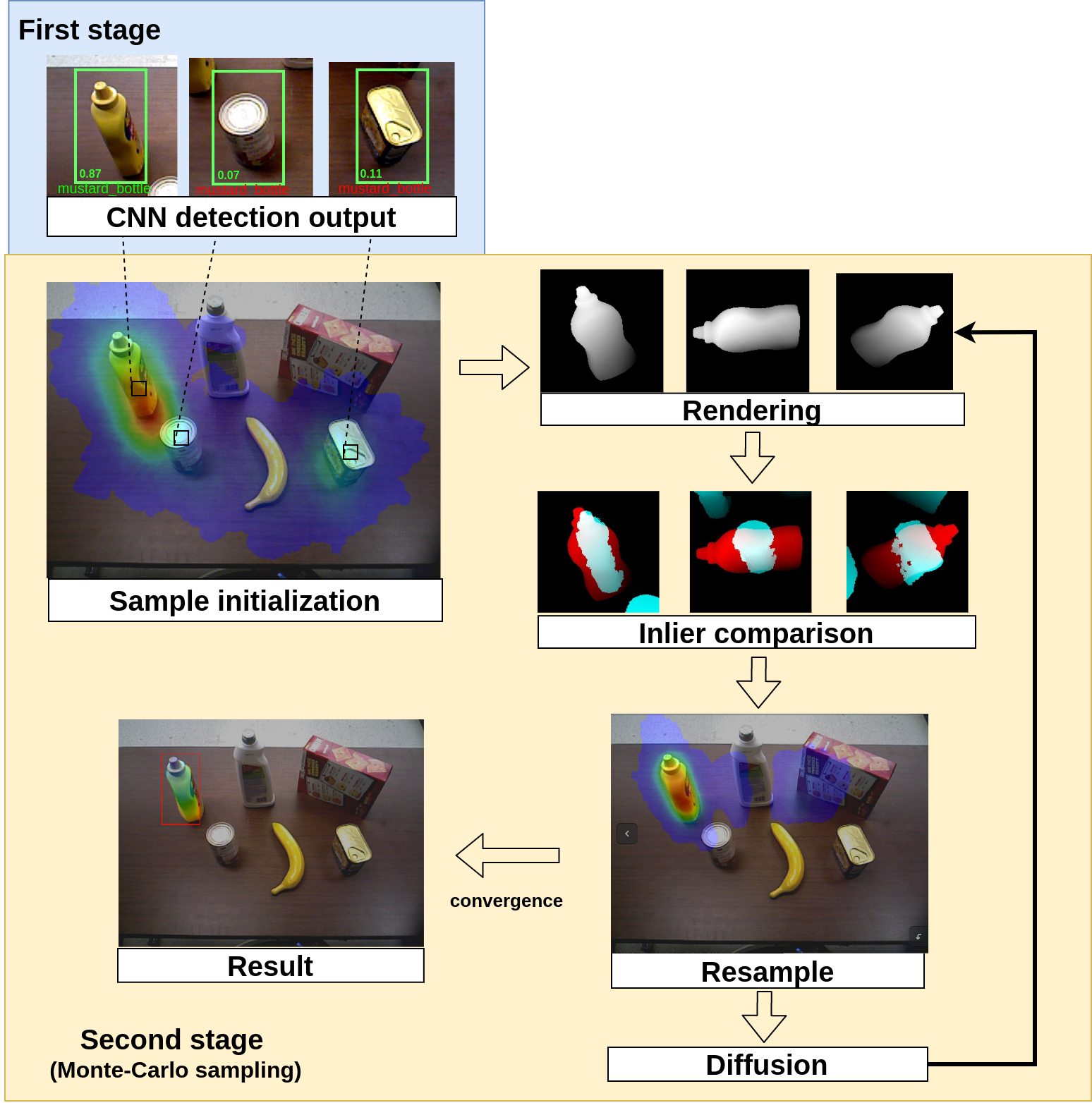}
\caption{Two-stage paradigm. $1^{st}$ stage uses CNN for object detection, $2^{nd}$ stage uses Monte-Carlo sampling to estimate object 6 DoF pose.}
%}
    \label{fig:two_stage_pipeline}
\end{figure}
%\vspace*{-2mm}

\begin{figure*}[tbh]
    \centering
    \includegraphics[width=0.68\linewidth]{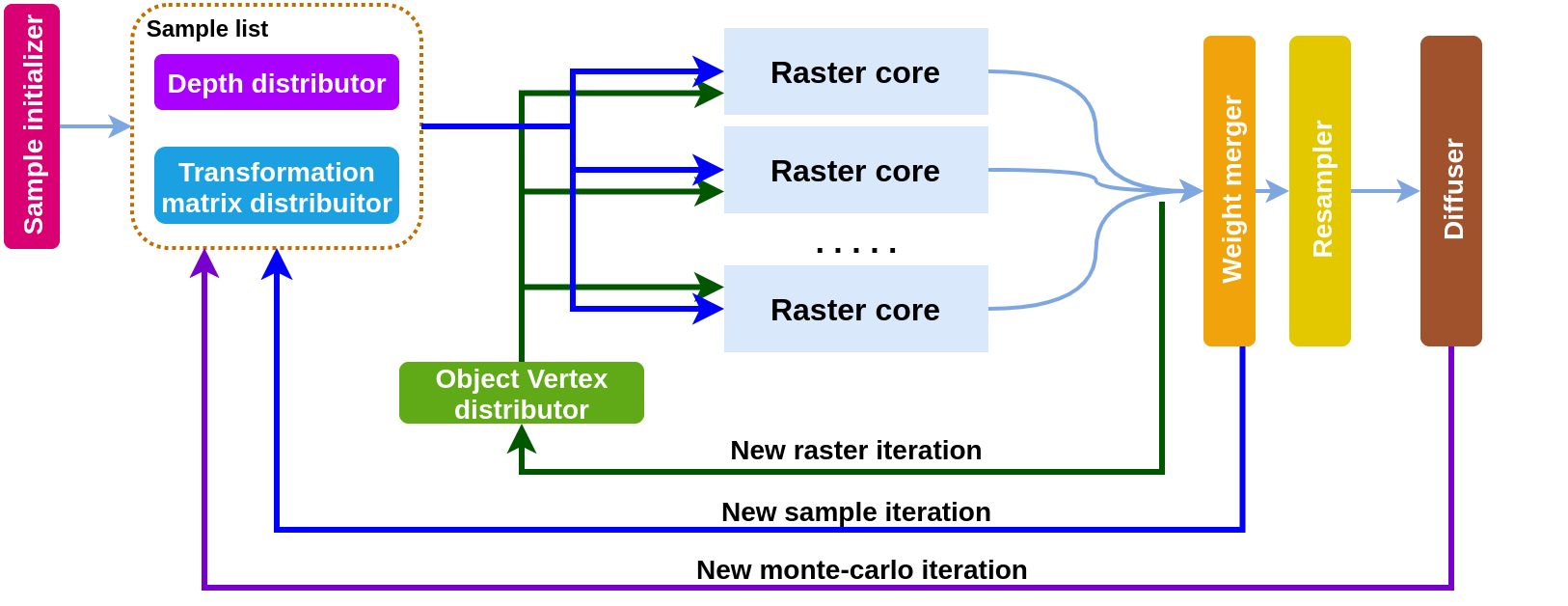}
    \caption{FPGA diagram: Depth and transformation distributor sends a new pose and depth region associated with each sample to each raster core. Object vertex distributor transfers each triangle of the geometric model to raster. Each raster core works in parallel to calculate inlier score of samples at different poses.
    }
    \label{fig:fpga_diagram}
    \vspace*{-4mm}
\end{figure*}

\section{Algorithm}
\label{sec:algo}
The two-stage paradigm for the discriminative-generative algorithm proposed in ~\cite{sui2018never} is shown in Fig.~\ref{fig:two_stage_pipeline}.  Our goal is to design an efficient hardware implementation of Monte-Carlo sampling used in the second stage of the algorithm for 6 DoF object pose estimation.
The input to this Monte-Carlo generative sampling algorithm 
is a series of bounding boxes around objects, with confidence scores and object class labels produced from any state-of-art object detection convolutional neural network (CNN).  This input represents the object probability distribution over the observed scene.
The CNN itself can be implemented by various network architectures such as VGG ~\cite{vgg16}, ResNet~\cite{resnet} and Squeezenet~\cite{iandola2016squeezenet} as discussed in~\cite{iccad}. However, the specific CNN architecture is not the focus of this paper.  Below we describe the generative sampling algorithm in detail, generally following the design presented in~\cite{sui2018never}.

%TThe goal for the first stage is to generate a probability distribution of the object class $o$ of possible locations over the entire image. In the first stage, the CNN outputs bounding boxes with their confidence scores and object class labels. Different from end-to-end object detection networks, this stage does not apply hard-threshold to the confidence score in order to avoid any false negatives. The CNN can be implemented by various network architectures such as VGG ~\cite{vgg16}, ResNet~\cite{resnet} and Squeezenet~\cite{iandola2016squeezenet} as discussed in~\cite{iccad}. However, the specific CNN architecture is not the focus of this paper.

%The goal of the second stage is to estimate object 6 DoF pose using a Monte Carlo generative sampling method. In the process, we generate object samples to represent our belief for the object pose and use iterated likelihood weighting to search for the optimal pose. In this way, we can recover the mistakes made by the first stage and achieve robust pose estimation.

Given an RGB-D observation ($Z_r$, $Z_d$) from a robot sensor $Z_r$ for RGB image and $Z_d$ for depth image, our goal is to maximize the conditional joint distribution $P(q,b|o, Z_r ,Z_d)$ for each object where $q$ is the 6 DoF pose for the object and $o$, $b$ are the class label and bounding box respectively from the CNN output.
%result of first stage. 
The problem can be formulated as:
\begin{align}
\label{eq: firstexp}
& P(q,b|o, Z_r ,Z_d)\\
\label{eq: secondexp}
& =P(q|b,o,Z_r,Z_d)P(b|o, Z_r ,Z_d)\\
& =\underbrace{P(q|b,o,Z_d)}_\text{pose estimation}\underbrace{P(b|o, Z_r)}_\text{detection}
\label{eq: objdist}
\end{align} 
%\textcolor{blue}{Note that the object detection is computed using only the RGB image while the pose estimation is computed using the depth image.}
The second-stage takes the object detection results from the CNN and performs Monte-Carlo sampling via iterative likelihood weighting. In the initial stage, the algorithm generates a set of weighted \textbf{samples} $\{q^{(i)}, w^{(i)}, b^{(i)}, z^{(i)}\}^M_{i=1}$ to represent the belief of the object pose over the entire image.  The value $q^{(i)}$ represents the 6 DoF pose of the sample object and $w^{(i)}$, $b^{(i)}$ and $z^{(i)}$ are associated with the probability, the bounding box of the object from the first stage, and the observed point cloud within the bounding box region, respectively. For each sample, given its object class $o$, pose $q^{(i)}$ and corresponding geometric model, the algorithm renders a 3D point cloud $r^{(i)}$ of the sample using z-buffering of a 3D graphics engine. The weight ${w^{(i)}}$ of each sample is updated to estimate how close the sample matches the observation. We use a pixel-wise \textit{inlier} function defined in Eqn.~\ref{eqn:inlier} to measure the matching between sample and observation:
\begin{equation}
\label{eqn:inlier}
    \text{Inlier}(p, p^{'}) = \mathbf{I}\left(||p-p^{'}||_2 < \epsilon\right),
\end{equation}
%\vspace*{-1mm}
where $p, p^{'}$ refers to a point in an observation point cloud \textbf{$z^{(i)}$} and a point in a rendered point cloud ${r^{(i)}}$ from the sample pose, respectively. $\mathbf{I}$ is the indicator function. An inlier is defined if a rendered point is within a certain distance threshold range $\epsilon$ from an observed point. The number of inliers is defined as: 
%in Eqn.~\eqref{eqn:inlier_num}:
\begin{equation}
\label{eqn:inlier_num}
    N^{(i)} = \sum_{a\in z^{(i)}} \text{Inlier}(r^{(i)}(a), z^{(i)}(a)),
\end{equation}
where $a$ is an index of a point within $z^{i}$. 
We can use this value to obtain two raw-pixel inlier ratios: $N^{(i)}/N_b$, where $N_b$ is the number of observation points within the bounding box $b^{(i)}$, and $N^{(i)}/N_r$, where $N_r$ is the number of rendered points within the bounding box.

Next, using these ratios and probability $c$ from the CNN, the weight $w^{i}$ for each sample is computed as:
%defined as a combination of probability $c$ from the first stage with raw-pixel inlier ratio: 
%in Eq.\ref{eqn:weight}.
\begin{equation}
\label{eqn:weight}
 w^{(i)} = \alpha * \frac{N^{(i)}}{N_b} + \beta * \frac{N^{(i)}}{N_r} + \gamma * c,
\end{equation}
%where $N_b$ is the number of observation points within the bounding box $b^{(i)}$, $N_r$ is the number of rendered points within the bounding box, \textcolor{red}{$c$ is the probability from the neural network }and 
where $\alpha$, $\beta$, $\gamma$ are the coefficients that are empirically determined and sum up to 1. 

To get the the optimal pose $q^*$, we follow the procedure of importance sampling~\cite{importance-sampling} to assign a new weight to each sample. During this process, each sample pose, $q^{(i)}$, is diffused with a Gaussian distribution in the space of 6 DoF poses with a small $\delta$ to increase sample variance:
%, as shown in Eqn.~\eqref{eqn:diffuse}: 
\begin{equation}
    \label{eqn:diffuse}
    q^{(i)} = (x,y,z, roll, pitch, yaw) + \mathcal{N}(0,\delta).
\end{equation}
%shown in Equation.~\eqref{eqn:diffuse} with a small $\delta$.}
Once the average sample weight is above a threshold, $\tau$, we consider the algorithm converged and $q^*$ will be selected as the sample with the highest weight $w^*$.

The most computationally expensive step in this process is the rendering and sample weight computation, which includes a pixel-wise inlier calculation. The amount of computation and memory grows linearly with the number of samples we choose for the design. Even though modern GPUs can achieve high parallelism, the high energy consumption makes them less suitable for mobile platforms such as autonomous robots.  In addition, their runtimes may still not allow for real-time operation. 
Our goal is to design various optimizations that can be implemented directly in hardware in order to achieve both faster runtime and reduced energy consumption. 
%We believe that through algorithm optimization, we can achieve the same speed performance while being energy- and memory-efficient for customized circuit design.

\section{Methodology}
\vspace{-1mm}

Our FPGA implementation of Monte-Carlo sampling is illustrated in Fig.~\ref{fig:fpga_diagram}.
The CNN object detection output consisting of probability and bounding box information is stored on off-chip memory and transferred to the second stage Monte-Carlo sampling module.
The \textit{sample initializer} generates $N$ samples (\textit{sample list} in Fig.~\ref{fig:fpga_diagram}) and the information for each sample (i.e., the 6DoF pose, bounding box region, and geometric model) is distributed to a \textit{raster core} through a \textit{transformation matrix distributor}, \textit{depth distributor} and \textit{object vertex distributor}. Each \textit{raster core} performs rasterization and inlier comparison on a single sample at a time.
% Since the number of samples $N$ is more than the number of raster cores $m$, we need more than one iteration to process all the samples (\textit{sample iteration} in Fig.\ref{fig:fpga_diagram}).
%In one \textit{sample iteration}, we distribute to each raster core in parallel: %\textit{\textbf{a)}} a transformation matrix based on the sample's associated pose %\textit{\textbf{b)}} a depth region corresponding to the sample's bounding box.
% \begin{enumerate}
%     \itemsep -.01in
%     \item transformation matrix of the sample's associated pose
%     \item depth region of the sample's bounding box.
% \end{enumerate}
% Each raster core is responsible for processing one sample. As shown in Fig.~\ref{fig:raster_core}, the raster core takes one triangle (defined by 3 vertices) at a time from the vertex distributor and sequentially operates on all the triangles in the object geometry model The raster core applies the transformation matrix and rasterizes the triangle to generate a depth map. Then it performs a pixel-wise depth comparison with the observation depth region to calculate an inlier score (\textit{raster iteration} in Fig.~\ref{fig:fpga_diagram}), as shown in Fig.~\ref{fig:render}.
The \textit{weight merger} step will fetch the inlier scores from each raster core, calculate the weights for each sample, and send them to the \textit{resampler}.
After all the samples are processed, the \textit{resampler} generates a new \textit{sample list} of 6 DoF poses based on the weight of each sample. Finally, the \textit{diffuser} stage adds Gaussian noise to each sample's 6 DoF pose. We then start a new \textit{Monte-Carlo iteration} for the new sample list.
% Beforestarting  the  nextMonte-Carlo  iterationa  Gaussian  noise  isadded  to  the  new  poses,  in  thediffuserstage,  to  have  morevariety among the samples.
We will next describe each of these steps in more detail in the following subsections.

 \begin{figure}[t]
    \centering \includegraphics[width=0.72\columnwidth]{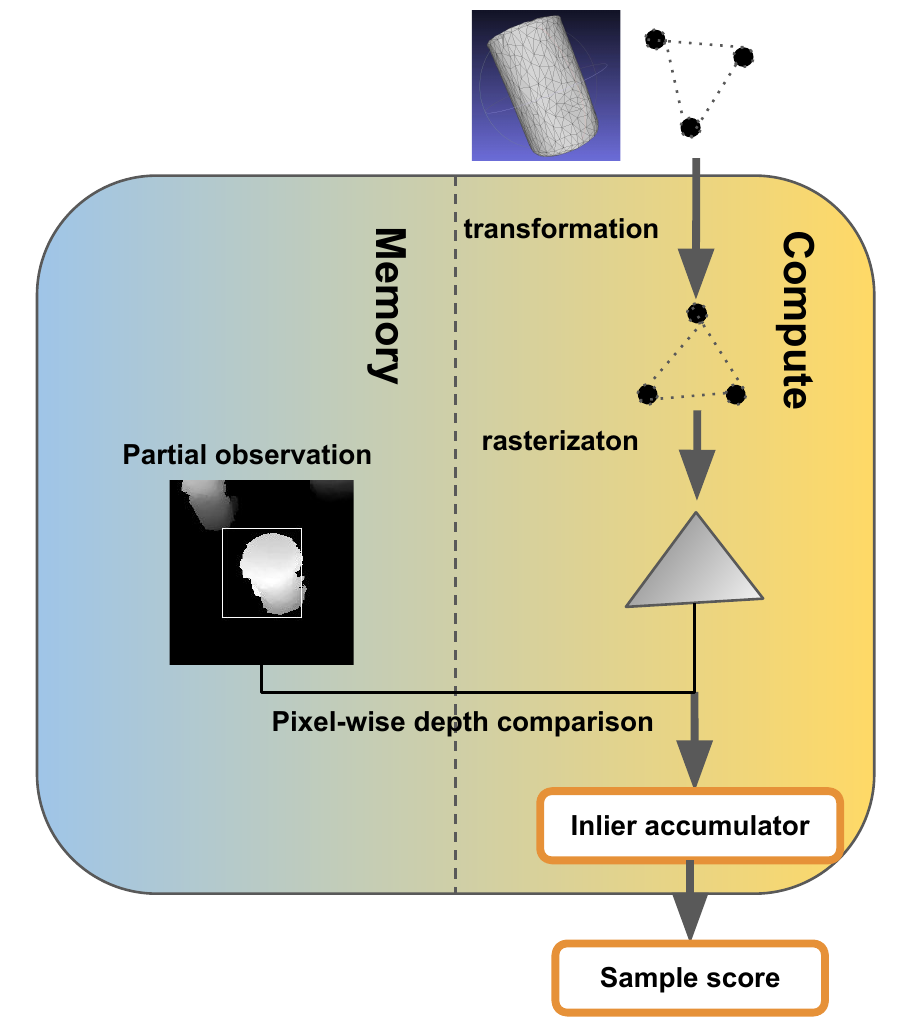}
    \vspace{-2mm}
    \caption{Raster core flow: Memory unit stores a region in the observation depth defined by sample bounding box. Raster core operates on a single triangle at a time.  The raster core pipelines the transformation, rasterization, and pixel-wise depth comparison between rasterized pixel and observation depth and outputs an inlier score.}
    \label{fig:raster_core}
    \vspace{-1mm}
\end{figure}

%We achieve computation efficiency by implementing fixed-point calculations for all operations in the Monte-Carlo sampling algorithm. In addition, we further improve the efficiency through a deep pipelined raster core, simplified inlier computation, parallel depth distribution, and efficient sorting and sampling. We detail these steps in the following subsections.

% {Raster core operates on all the triangle meshes in the object geometry model and all the steps inside the raster core is pipelined to increase the throughput.}
\vspace{-1mm}
\subsection{Rasterizing}
\vspace{-1mm}
Rasterization is a  process in computer graphics that converts a geometric model defined by vertices and faces to a raster image, defined by a series of pixels each with a depth value. The result of rasterization is an image of what a 3D object would look like at a certain view point. To implement rasterization in hardware, we designed a specialized \textit{raster core} processing unit that pipelines the rasterization and inlier comparison steps for a given sample. The processing unit is illustrated in  Fig.~\ref{fig:raster_core}.
At every \textit{raster iteration}, a triangle from the geometric model is transformed with a sample's 6 DoF pose and rasterized, after which a depth value at each pixel within the triangle is calculated and compared to the observed depth region stored in the raster core.
The comparison results are accumulated and output as an inlier score after all the triangles within the geometric model are processed. By pipelining these two steps, there is no need to store the rasterization result, and instead we only keep track of the inlier score from each sample.
%and we don't stall the rasterization process during inlier computation.}

\begin{figure}[bht]
\vspace*{-4mm}
\centering
\subfloat[full raster]{
\label{fig:mdleft}
{\includegraphics[width=0.26\linewidth]{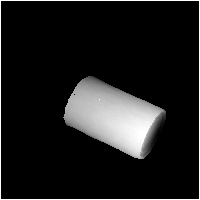}}}\hfill
\subfloat[partial raster 1]{\label{fig:mdcenter}{\includegraphics[width=0.26\linewidth]{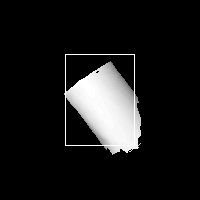}}}\hfill
\subfloat[partial raster 2]{\label{fig:mdright}{\includegraphics[width=0.26\linewidth]{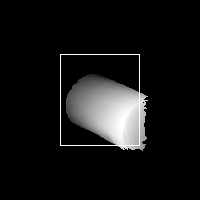}}}\hfill
% \subfloat[My second picture]{\label{fig:mdright}{\includegraphics[width=0.25\linewidth]{figures/partial_3.png}}}
\caption{Partial rendering: (a) full rasterization of the object, (b) and (c)  partial rasterizations within the sample bounding boxes.}
\label{fig:full_partial}
\end{figure}

We further reduce computational complexity and memory utilization by using partial rasterization. Note that since Eqn.~\eqref{eqn:weight} only pertains to the region within the bounding box, we only need to rasterize within this region.  This partial rasterization is illustrated in Fig.~\ref{fig:full_partial}.
Backface culling is a standard algorithm inside a 3D graphic pipeline that removes the faces of the object model occluded by some other triangle~\cite{culling}. For our purposes, we apply backface culling to reduce the total number of rasterized triangles by using the dot product between the surface normal and the camera point of view direction to judge if a face is occluded.
On average, we found that we can reduce the number of rendered triangles by approximately 50\% using this technique.
%As shown in Fig.~\ref{fig:BF_PERF}, with backface culling, on average the number of rendered triangles is reduced by approximately $50\%$.

%In each raster core, we only store the depth region associated with the sample.
% \begin{figure}[t]
% \centering
% \subfigure[full raster]{\includegraphics[width = 2.5cm]{figures/full_raster.png}
% \subfigure[partial raster 1]{\includegraphics[width = 2.5cm]{figures/partial_1.png}
% \subfigure[partial raster 2]{\includegraphics[width = 2.5cm]{figures/partial_2.png}
% \caption{\textcolor{red}{Visualization of partial rasterization for a soup can. (a) full raster of the object. (b),(c) partial raster of the soup can within the bounding box.}}
% \label{fig:render}
% \end{figure}

%\begin{figure}[t]
%    \centering
%    \includegraphics[width=0.7\linewidth]{figures/INL_BF_CULLING.png}
%    \caption{Backface culling. The orange line shows the number of triangles in the model and the green one shows the average number of rastered triangles with backface culling applied.}
%    \label{fig:BF_PERF}
%\end{figure}
\vspace*{-3mm}
\subsection{Inlier}
In the original algorithm described in \cite{sui2018never}, a 3D point cloud is used to represent the rasterized sample and observation. In order to reduce the computation and memory overhead, we wanted to modify the inlier comparison to a 1D depth representation. Given a point ($x, y, z$) in an observation pointcloud and a point($x', y',z'$) in a rendered point cloud the 3D Euclidean distance between two points can be computed noting that:
%\vspace*{-2mm}
 \begin{equation}
% \begin{align}
     d = \sqrt{(x-x')^2 + (y-y')^2 + (z-z')^2}.\\
%\end{align}
\label{eqn:3d_dist}
 \end{equation}
 Given a depth $z$ at pixel ($p_x$,$p_y$), the $x$ and $y$ values can be calculated as:
\vspace*{-2mm}
 \begin{equation}
 \begin{split}
     x = (p_x - C_x)\cdot z / f_x \\
     y = (p_y - C_y)\cdot z / f_y
\label{eqn:3d_depth}
\end{split}
 \end{equation}
where $C_x$, $C_y$, $f_x$. $f_y$ are camera intrinsic parameters (i.e., center offset and focal length).
By substituting the $x$, $y$ values in Eqn.~\eqref{eqn:3d_dist} with the formulation in Eqn.~\eqref{eqn:3d_depth}, we see that for the same pixel ($p_x$, $p_y$) the distance differences in the $x$ and $y$ directions are proportional to the distance differences in the $z$ direction.  That is:
\begin{equation}
\begin{split}
     x-x' &= (p_x - C_x)\cdot z / f_x - (p_x - C_x)\times z' / f_x  \\
      &= (p_x - C_x)\cdot  (z - z') / f_x  \\
      &\propto  (z-z')  \\
     y-y' &= (p_y - C_y)\cdot  z / f_y - (p_y - C_y)\cdot z' / f_y  \\
     &= (p_y - C_y)\cdot  (z - z') / f_y\\
     &\propto  (z-z')
\label{eqn:3d_euc}
\end{split}
\end{equation}
Therefore, we can approximate the 3D Euclidean distance computation with a much simpler 1D depth comparison without affecting the pose estimation accuracy.

% Fixed-point quantization is carefully designed for all operations in our design while preserving computation precision. Fixed-point operation reduces hardware resources for computation compared to floating-point and can be accelerated using integer arithmetic. It also saves memory for storage.

\subsection{Depth Distributor}

\begin{figure}[t]
    \centering
    \includegraphics[width=.75\columnwidth]{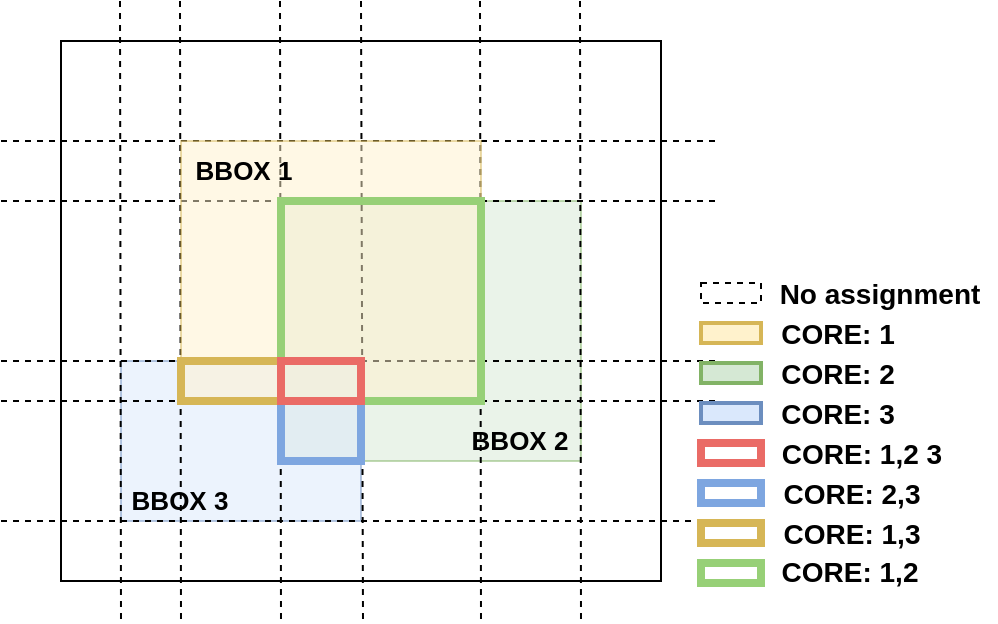}
    \caption{Illustration of depth distribution. Each region shown within dotted lines is distributed to a raster core and overlapping regions (highlighted by the different colored boxes) will be distributed in parallel to multiple raster cores.}
    \label{fig:depth_distri}
\end{figure}

% Inside the algorithm each core needs to compare the rendered object with the corresponding depth and the object model for each of them is always the same.
In the inlier calculation step, each rendered sample is compared with its corresponding observation depth region defined by a bounding box generated from the CNN output in the first stage. Therefore, each raster core must read a depth region from the entire depth image stored in on-board memory. Naively, if we distribute each region to each raster core in series, we need to repeatedly access the same memory location multiple times for the overlapping areas. Instead, we designed a depth distributor, as illustrated in Fig.~\ref{fig:depth_distri}, to reduce the amount of redundant memory accesses for overlapping depth regions. Our algorithm first divides the depth image into multiple sub-regions and identifies each region with a corresponding raster core number. The data from the overlapping regions are then read from memory once and distributed to multiple raster cores in parallel. In this way, the more overlapping areas we have among different regions, the faster the depth distribution can be completed. In particular, assuming that as more \textit{Monte-Carlo iterations} are completed more samples will converge to the same bounding box, the runtime and power consumption of the depth distributor for later iterations will decrease.

\subsection{Sorting and Resampling}

% From the start array we pick the information of the particles needed to evaluate the weights, while in the end one we insert the most likely samples' information with a slight distortion.
%\textcolor{green}{We can decide to execute the resampling stage in the entire set of current samples or just a smaller part, like the top 100. In the latter case the sample must be  sorted according to the corresponding weight. }

To execute the resampling stage, we sort the samples by their corresponding weight.  While sorting is not strictly required for importance sampling, we can sort the samples by their weights and only resample from the top $x$\% to further reduce resampling memory accesses.
Sorting a sample generally requires moving all its object pose information (i.e., \textit{x, y, z, roll, pitch, yaw}). %and the Heatmap index,  according to the values generated by the rendering core. Furthermore, we are interest to copy only the informations of the resampled particles, which will be used in the next iteration,  instead of all of them.
However, to reduce data movement, we sort the index of the sample based only on the sample weight. %give an identification number to all the unsorted particles, called relative index starting from $0$ up to $N-1$.
%This sorted index is stored in a separated array and sorted as previously described.
Once the sorting step is complete, we conduct sampling to generate new sample indices.

\begin{figure}[t]
    \centering
    \includegraphics[width=.75\linewidth]{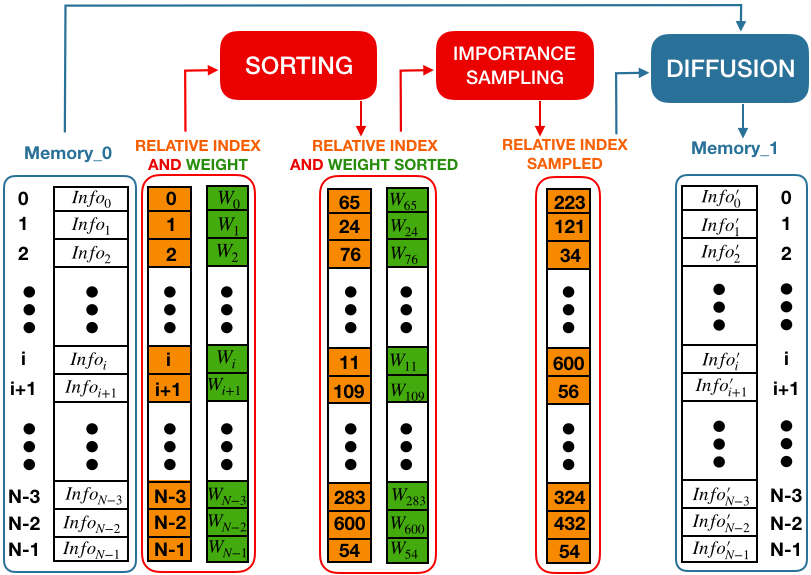}
    \caption{Sorting and resampling. Starting with our $N$-entry \textit{sample list} in $Memory\_0$, we sort the entries by weight in ascending order.  In this example, $W_{65} > W_{24}> W_{76}$, etc. $Memory\_0$ and $Memory\_1$ will be read and written alternatively across each iteration.}
    \label{fig:Resampling}
\end{figure}

%The latter is used to read the informations from the start memory, diffuse and write it to the ending memory.

Recall from Section~\ref{sec:algo} that resampling uses importance sampling to generate new samples from the sample weight distribution. We store the cumulative density function (CDF) of the normalized sample weight $x$ in an array, $\Phi(x)$.
A random number $r \in U(0,1)$ is generated and compared with values in the CDF array until we find the first sample $i$ where $r < \Phi(x^{(i)})$.
%\textcolor{blue}{A random number $r \in U(0,1)$ is compared with all the values in the CDF array. We find the first sample $i$ where $r < \Phi(x^{(i)})$.}
The number of memory reads to the CDF array increases as $r \rightarrow 1$, since the memory locations must be analyzed starting from the zero cell up to the desired one.
To reduce memory accesses, we use a separate memory to store a set of threshold values $\{t_0, t_1, ..., t_n\}$ with a constant step, where $t_i$ $\in$ [0,1] such that we first execute a coarse-grained search to find region [$t_k$, $t_{k+1}$] where $r$ falls, and then do a fine-grained search for $i$ where $\Phi(x^{(i)})$ $\in$ [$t_k$, $t_{k+1}$].
From our experiments, we found that this technique greatly reduces the average number of memory accesses from  410 to 10. %as shown in in table ~\ref{tab:THR_NO_YES}. }

%is possible to proceed to the cumulative normalised sum of the sample's weights and start the resampling. The latter core instead of copying the particle informations just copy the resampled relative index in S_SAND-a temporary memory.
%During each Monte-carlo iteration, we implement a ping-pong buffer,
To further speed up execution time, we implemented a ping-pong buffer for the diffusion stage, as illustrated in Fig.~\ref{fig:Resampling}.  We alternate fetching sample information from either $Memory$\_$0$ or $Memory$\_$1$ using the new sample indices generated from the resampler, add Gaussian noise to the 6 DoF pose, and save the new samples in the opposite memory buffer.
%pre-comparison between the random number $r$ and the thresholds $t$ to select a smaller range for memory read. This greatly reduces the memory access and the result is shown in Fig.~\ref{fig:THR_NO_YES}.

% \begin{table}[!hbt]
%     \centering
%     \begin{tabular}{|c|c|c|c|} \hline
%  Iteration   & \# read without & \# read  with & speed up \\ \hline
%  1 & 423.85 & 9.43 & 44.93 \\
%  2 & 410.23  & 9.30  & 44.095 \\
%  3 & 401.86 &9.36  & 42.95 \\
%  4 & 402.51 & 9.45 & 42.48 \\ \hline
%     \end{tabular}
%     \caption{Caption}
%     \label{tab:my_label}
% \end{table}

\section{Experimental Results}
We implemented our Monte-Carlo sampling algorithm
on a Xilinx\textsuperscript{\textregistered} Virtex UltraScale FPGA ZCU102 board using the Vivado HLS high-level synthesis tool.  Given the memory and computational resources of this board, we can fit a total of 20 raster cores in our design.
In general, the more samples we use, the better we can approximate sample distribution. In our case, we chose to process a total of 620 samples because it is sufficient to describe our algorithm search space, so 31 \textit{sample iterations} are needed to render all the samples.
%For our experimental setup, we implemented a 20 raster-core design on a Xilinx\textsuperscript{\textregistered} Virtex UltraScale FPGA ZCU102 board using the Vivado HLS high-level synthesis tool.  In general, the more samples we use, the better we can approximate sample distribution. \textcolor{red}{In our case, we chose to process a total of 620 samples because it is sufficient to describe our algorithm search space. }

We compared  runtime, power, energy consumption, and accuracy of our FPGA implementation to a CPU-GPU hybrid reference design implemented on two platforms: 1) an Nvidia\textsuperscript{\textregistered} Titan Xp with and Intel Xeon E5, and 2) an Nivida\textsuperscript{\textregistered} Jetson TX2 with a quad-core ARM A57. We note that both GPU platforms are more powerful than our FPGA in terms of memory capacity, compute resources, and clock frequency.
Sample initialization, resampling, and diffusion are done on the CPU since their operations are sequential in nature, while sample rendering and inlier computation is done on the GPU. We use OpenGL to render all the samples and program the CUDA cores to perform the inlier computation. We create a kernel where every pixel distance comparison is assigned to a CUDA thread and processed concurrently. Since the GPU has high memory bandwidth, we keep one copy of the observation depth in the GPU memory such that each sample accesses the observation depth to compute the inlier.

\begin{figure}[ht]
\vspace{-2mm}
\centering
\subfloat[RGB image]{
\label{fig:rgb}
{\includegraphics[width=0.45\linewidth]{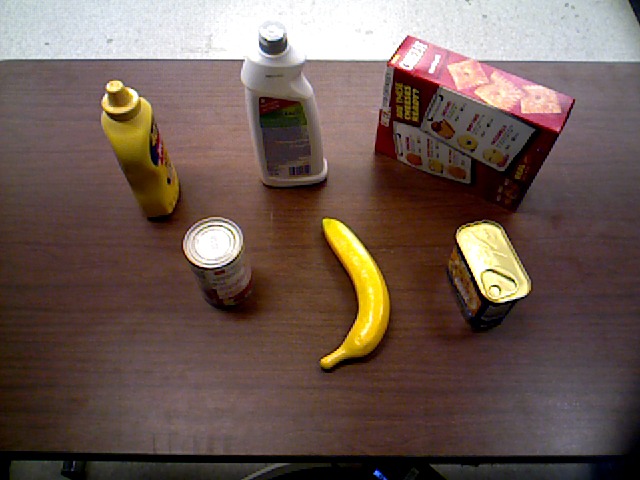}}}\hfill
\subfloat[Depth image without table.]{\label{fig:depth}{\includegraphics[width=0.45\linewidth]{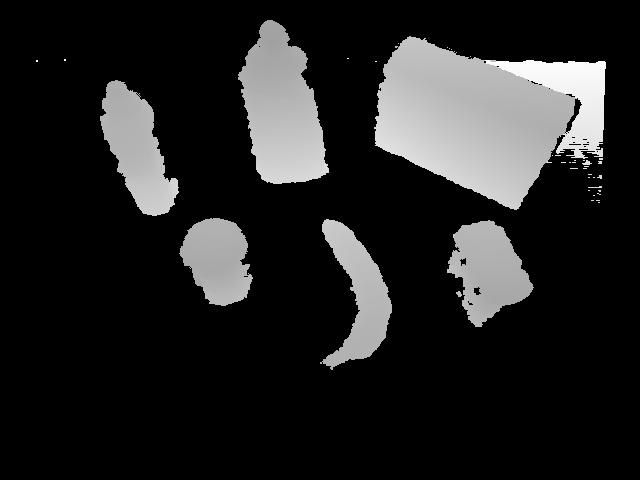}}}\hfill
% \subfloat[My second picture]{\label{fig:mdright}{\includegraphics[width=0.25\linewidth]{figures/partial_3.png}}}
\caption{A test scene containing objects from the YCB dataset.}
\label{fig:scene}
\end{figure}

The dataset used in the experiments contains scenes collected by a Kinect RGBD camera with objects from the YCB dataset \cite{posecnn}. Each scene captures a depth image of size 640 X 480. In each scene, 5--7 different objects are placed on a table.  An example scene is shown in Fig.~\ref{fig:scene}. We choose to test these 5 different objects for their different sizes and symmetries.

\subsection{Runtime}
Table~\ref{tab:runtime} reports the average runtime for the rendering and inlier stages on both FPGA and GPU platforms.
%We take the average per iteration runtime of 50 iterations of the algorithm runs. The runtime is shown in Table.~\ref{tab:runtime}.
Note that the FPGA implementation has a faster runtime than the Jetson version and slightly slower runtime compared with Titan.
%The Titan Xp has 12GB on board DDR5 memory and 3840 CUDA cores, which makes it able to render all the samples in parallel. The inlier computation can also be done in parallel for each pixel within each sample rendered pointcloud.

\begin{figure}[t]
\vspace*{-2mm}
    \centering
    \includegraphics[width=\linewidth]{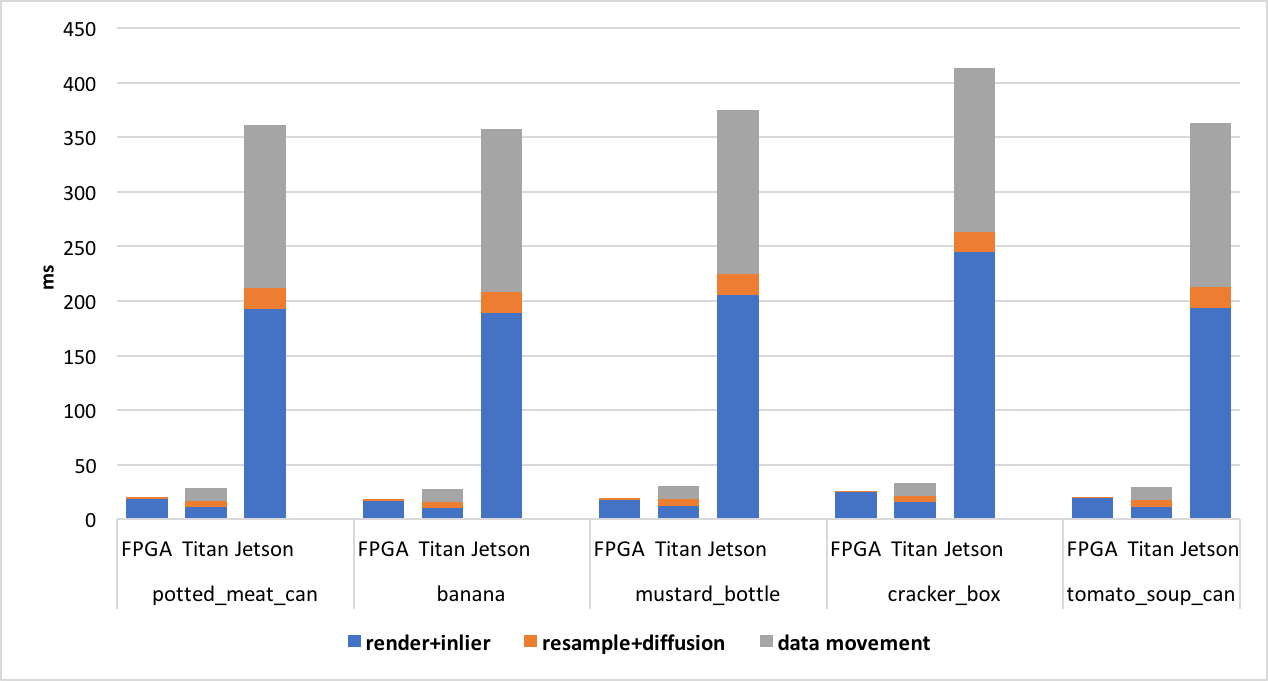}
    \caption{Average runtimes for one Monte-Carlo iteration.}
    \label{fig:runtime_benchmark}
\end{figure}

In Fig.~\ref{fig:runtime_benchmark} we show the average runtime per entire \textit{Monte-Carlo iteration}, broken down into three stages:  1) rendering+inlier computation, 2) data transfer between GPU and CPU, and 3) resampling and diffusion computation.  Since our FPGA implementation keeps the entire Monte-Carlo sampling algorithm on board, the total data transfer time is greatly reduced and thus has advantages for per-iteration runtime.

%Note that while Monte-Carlo inference processes each object in series, the robot can start object manipulation as soon as the first object completes. On average, it takes 50 \textit{Monte-Carlo iterations} for the algorithm to converge for each object. Given the average runtimes from Table~\ref{tab:runtime}, our FPGA implementation can process a single object in approximately 1 second, which can be considered real-time performance.

Note that while Monte-Carlo inference processes each object in series, the robot can start object manipulation as part of a pick-and-place action as soon as the first object completes.
On average, it takes 50 \textit{Monte-Carlo iterations} for the algorithm to converge for each object. Given the average runtimes from Table~\ref{tab:runtime}, our FPGA implementation can process a single object in approximately 1 second.
Since a robot movement can take a few seconds to complete, we can start to pick the next object without stalling the robot action; thus, we consider 1 second as real time processing for this application.

\begin{figure}[t]
\vspace*{-6mm}
    \centering
    \includegraphics[width=0.75\linewidth]{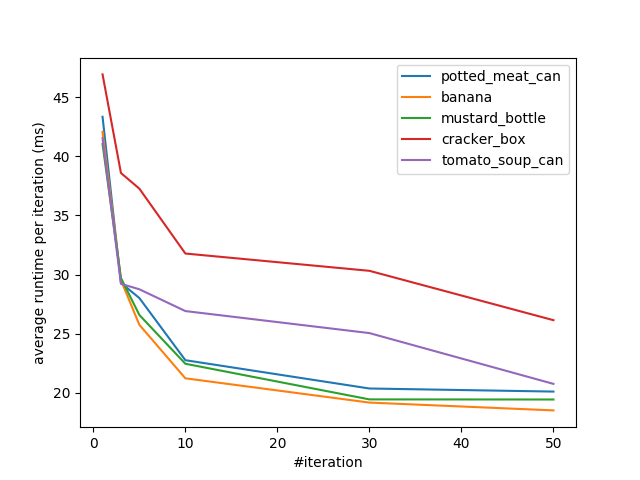}
    \vspace*{-2mm}
    \caption{FPGA per Monte-Carlo iteration runtime vs.\/ iteration count.}
    \label{fig:50_iter_time}
\end{figure}

Finally, the benefit of our depth distribution implementation  is shown in Fig.~\ref{fig:50_iter_time}.
Here we chose to test average runtime per iteration for 50 iterations.
Note that the average runtime decreases over the iterations as the objects converge.

\begin{table}[t]
    \centering
    \resizebox{\columnwidth}{!}{\begin{tabular}{|L||L|L|L|L|L|}
    \hline
     &banana & cracker box & potted meat can & mustard bottle & tomato soup can \\
    \hline \hline
        % FPGA & \textbf{26.72ms} & 48.08ms & 31.897ms & \textbf{30.22ms} & \textbf{31.85ms} \\
        FPGA & 17.08ms & 24.70ms & 18.67ms & 18.01ms & 19.32ms \\
        \hline
         Titan & 10.24ms & 15.78ms & 11.29ms & 12.52ms & 11.66 ms \\
         \hline
         Jetson & 188.81ms & 244.61ms & 192.45ms & 205.53ms & 193.66ms\\
         \hline
    \end{tabular}}
    \caption{Average runtimes for render+inlier process: FPGA at \textbf{200MHz}, Titan Xp at \textbf{1.4GHz} and Jetson TX2 at \textbf{1.3GHz}}.
    \label{tab:runtime}
\end{table}

%\textbf{Note also that the GPU cannot take advantage of this depth distribution optimization.}

% \begin{figure}[t]
%     \centering
%     \includegraphics[width=0.9\linewidth]{figures/stacked_runtime.png}
%     \caption{Runtime distribution for each stages}
%     \label{fig:my_label}
% \end{figure}
\vspace{-1mm}
\subsection{Resource and Power}
\vspace{-1mm}
The resource utilization for each stage of our algorithm is shown in Table~\ref{tab:resource} and the power and energy consumption of each implementation is shown in Table~\ref{tab:power}. The FPGA power is collected from the Vivado\textsuperscript{\textregistered} power analyzer, while the Titan power is estimated through the Nvidia\textsuperscript{\textregistered} Management Library, and Jetson power is measured through an on-board power monitor.
Note that for our FPGA implementation, resampling and diffusion steps are responsible for less than 13\% of total power/energy.
In addition, the Titan and Jetson GPU only performs rendering and inlier computations and not resampling and diffusion stages since these are on the CPU.
Just focusing on rendering+inlier power and energy, we see that the FPGA implementation is 33X more power efficient and 21X more energy efficient than the Titan Xp. Compared against the low power Jetson, our FPGA implementation is slightly more power efficient and 12X more energy efficient.
Moreover, our solution is scalable to any FPGA platform by choosing the number of raster cores to balance the runtime versus resource and power tradeoff.

%The energy is calculated by taking average power $\times$ average render+inlier runtime, since only rendering and inlier are performed on the GPU. Our FPGA implementation is 28X more power efficient and 18X more energy efficient than the Titan Xp implementation. While our FPGA implementation achieves similar power efficiency as the Jetson, it is 10X more energy efficient due to its faster runtime.
%Note that for the GPU implementations, power and energy is measured only for the rendering and inlier computation, and not the data transfer or resampling and diffusion stages since these are on the CPU.  Therefore, our reported power/energy improvements on the FPGA are lowerbounds; if we consider the entire Monte-Carlo sampling process, our FPGA implementation would show even greater power/energy advantages.

\begin{table}[ht]
    \centering
    \begin{tabular}{|c||c|c|c|c|}
    \hline
    & BRAM36s & DSP48E2 & LUT & FFs \\
    \hline \hline
    sample initializer & 0 & 6 & 2123 & 2409\\
    \hline
    20 raster cores & 480  &  920  & 146000 & 172780  \\
    \hline
    resample  & 1 & 10 & 926 & 855 \\
    \hline
    diffuse  & 0.5  &  13 & 3588&1990 \\
    \hline
    \end{tabular}
    \caption{Resource utilization for each stage }
    \label{tab:resource}
\end{table}
\vspace*{-2mm}

\begin{table}[h!]
    \centering
    \begin{tabular}{|c||c|c|c|c|}
    \hline
%         & FPGA(200MHz) & Titan(1.4GHz) & \textcolor{red}{Jetson(1.3GHz)} \\
         & FPGA & FPGA  & Titan & Jetson \\
         & whole flow & render+inlier  & render+inlier  & render+inlier  \\
         \hline \hline
        $P_{ave}$ & 3.85W & 3.36W & 110.34W  & 3.78W\\
        \hline
        $E_{ave}$ & 75.32mJ & 65.70mJ & 1357.12mJ & 775.62mJ\\
         \hline
    \end{tabular}
    \caption{Average power and energy consumption}
    \label{tab:power}
\end{table}
\vspace{-2mm}
\subsection{Accuracy}
\vspace{-1mm}
To evaluate the accuracy of our implementation, we chose 5 objects and 9 different scenes, where each object occurs 5 times in the scene. We ran our algorithm 5 times for each scene to avoid randomness in the result. We use the average distance metrics ADD and ADD-S as defined in~\cite{posecnn} to calculate the point distance error between predicted pose and ground truth pose for symmetric and non-symmetric objects respectively. Both GPU and FPGA implementations achieve around 52\% pose estimation accuracy under an ADD threshold of 4 cm.
We see that even though we simplified the inlier calculation and rasterization steps, the FPGA implementation achieves similar accuracy as the GPU implementation.
%and we calculate the point-distance error of the object at the predicted pose versus the ground truth pose.
% \begin{figure}[t]
%     \centering
%     \includegraphics[width=0.75\linewidth]{figures/accuracy.png}
%     \caption{\textcolor{red}{Overall pose estimation accuracy of 5 YCB objects. The x-axis is the Average Distance Difference(ADD) between the predicted pose from Monte-Carlo inference and the ground-truth pose. The y axis is the ratio of the number of predicted pose within certain ADD threshold over all the predicted poses}}
%     \label{fig:accuracy}
% \end{figure}

\section{conclusions}

%\textbf{Based on the final design is possible to state that  the power consumption of the FPGA makes it suitable for embedded systems, differently from the GPU.  Furthermore,  The module  provides a similar run time even if the number of cores is limited and the clock frequency is seven times lower than the GPU. 
%As a final consideration the flexibility in terms of number of cores, without the degradation of the performances, gives the possibility  to use the FPGA to accelerate other algorithms.  }
In this paper, we have shown an effective hardware implementation of a Monte-Carlo sampling algorithm, as part of the two-stage discriminative-generative method used for pose estimation for robot manipulation. With our FPGA implementation, we are able to achieve real time performance with significantly reduced energy consumption, compared to either a high-performance or low power GPU implementation.
%Even though our FPGA platform is much more constrained in terms of memory and computation resources and has slower clock frequency, we achieve approximately the same performance.  Furthermore, we obtain substantial saving in terms of power and energy consumption which makes our implementation more suitable for mobile robots. 
Future work will consider adding a deep pipelined feature extraction step, along with rasterization, to provide higher accuracy for objects pose estimation in more clustered environments.
\newpage

\bibliographystyle{unsrt}
\balance
\bibliography{ref}

\begin{thebibliography}{10}

\bibitem{srivastava2014dropout}
Nitish Srivastava, Geoffrey Hinton, Alex Krizhevsky, Ilya Sutskever, and Ruslan
  Salakhutdinov.
\newblock Dropout: A simple way to prevent neural networks from overfitting.
\newblock {\em Journal of Machine Learning Research}, 15(1):1929–1958,
  January 2014.
\newblock doi:10.5555/2627435.2670313.

\bibitem{goodfellow2014explaining}
Ian~J. Goodfellow, Jonathon Shlens, and Christian Szegedy.
\newblock Explaining and harnessing adversarial examples.
\newblock {\em CoRR}, abs/1412.6572, 2015.

\bibitem{eykholt2018robust}
Kevin Eykholt, Ivan Evtimov, Earlence Fernandes, Bo~Li, Amir Rahmati, Chaowei
  Xiao, Atul Prakash, Tadayoshi Kohno, and Dawn Song.
\newblock Robust physical-world attacks on deep learning visual classification.
\newblock In {\em Proceedings of the IEEE Conference on Computer Vision and
  Pattern Recognition}, pages 1625--1634, 2018.
\newblock doi:10.1109/CVPR.2018.00175.

\bibitem{sui2017sum}
Zhiqiang Sui, Zheming Zhou, Zhen Zeng, and Odest~Chadwicke Jenkins.
\newblock {SUM}: Sequential scene understanding and manipulation.
\newblock In {\em IEEE/RSJ International Conference on Intelligent Robots and
  Systems (IROS)}, pages 3281--3288. IEEE, 2017.
\newblock doi: 10.1109/IROS.2017.8206164.

\bibitem{narayanan2016discriminatively}
Venkatraman Narayanan and Maxim Likhachev.
\newblock Discriminatively-guided deliberative perception for pose estimation
  of multiple 3d object instances.
\newblock In {\em Robotics: Science and Systems}, 2016.
\newblock doi:10.15607/RSS.2016.XII.023.

\bibitem{grip}
Xiaotong Chen, Rui Chen, Zhiqiang Sui, Zhefan Ye, Yanqi Liu, R.~Iris Bahar, and
  Odest~Chadwicke Jenkins.
\newblock {GRIP}: Generative robust inference and perception for semantic robot
  manipulation in adversarial environments.
\newblock {\em IEEE/RSJ International Conference on Intelligent Robots and
  Systems (IROS)}, 2019.
\newblock doi:10.1145/3240765.3243493.

\bibitem{sui2018never}
Zhiqiang Sui, Zhefan Ye, and Odest~Chadwicke Jenkins.
\newblock Never mind the bounding boxes, here's the {SAND} filters.
\newblock {\em arXiv:1808.04969}, 2018.

\bibitem{posecnn}
Yu~Xiang, Tanner Schmidt, Venkatraman Narayanan, and Dieter Fox.
\newblock {PoseCNN}: A convolutional neural network for 6d object pose
  estimation in cluttered scenes.
\newblock {\em Robotics: Science and Systems XIV}, 2018.
\newblock doi:10.15607/RSS.2018.XIV.019.

\bibitem{dope}
Jonathan Tremblay, Thang To, Balakumar Sundaralingam, Yu~Xiang, Dieter Fox, and
  Stan Birchfield.
\newblock Deep object pose estimation for semantic robotic grasping of
  household objects.
\newblock {\em arXiv:1809.10790}, 2018.

\bibitem{densefusion}
Chen {Wang}, Danfei {Xu}, Yuke {Zhu}, Roberto {Martín-Martín}, Cewu {Lu},
  Li~{Fei-Fei}, and Silvio {Savarese}.
\newblock Densefusion: {6D} object pose estimation by iterative dense fusion.
\newblock In {\em 2019 IEEE/CVF Conference on Computer Vision and Pattern
  Recognition (CVPR)}, pages 3338--3347, 2019.
\newblock doi:10.1109/CVPR.2019.00346.

\bibitem{sensorfusion}
László {Schäffer}, Zoltán {Kincses}, and Szilveszter {Pletl}.
\newblock A real-time pose estimation algorithm based on {FPGA} and sensor
  fusion.
\newblock In {\em International Symposium on Intelligent Systems and
  Informatics (SISY)}, Sep. 2018.
\newblock doi:10.1109/SISY.2018.8524610.

\bibitem{schaeferling2012object}
Michael Schaeferling, Ulrich Hornung, and Gundolf Kiefer.
\newblock Object recognition and pose estimation on embedded hardware:
  Surf-based system designs accelerated by {FPGA} logic.
\newblock {\em International Journal of Reconfigurable Computing}, 2012:6,
  2012.
\newblock doi:10.1155/2012/368351.

\bibitem{konomura2016fpga}
Ryo Konomura and Koichi Hori.
\newblock {FPGA}-based {6-DoF} pose estimation with a monocular camera using
  non co-planer marker and application on micro quadcopter.
\newblock In {\em 2016 IEEE/RSJ International Conference on Intelligent Robots
  and Systems (IROS)}, pages 4250--4257, 2016.
\newblock doi: 10.1109/IROS.2016.7759626.

\bibitem{fpga-icp}
Atsutake {Kosuge}, Keisuke {Yamamoto}, Yukinori {Akamine}, Taizo {Yamawaki},
  and Takashi {Oshima}.
\newblock A 4.8x faster {FPGA}-based iterative closest point accelerator for
  object pose estimation of picking robot applications.
\newblock In {\em IEEE 27th Annual International Symposium on
  Field-Programmable Custom Computing Machines (FCCM)}, pages 331--331, April
  2019.
\newblock doi:10.1109/FCCM.2019.00072.

\bibitem{murai2019visual}
Riku Murai, Paul Kelly, Sajad Saeedi, and Andrew Davison.
\newblock Visual odometry using a focal-plane sensor-processor.
\newblock 2019.

\bibitem{10.1007/978-3-319-16214-0_8}
Fynn Schwiegelshohn, Eugen Ossovski, and Michael H{\"u}bner.
\newblock A fully parallel particle filter architecture for {FPGAs}.
\newblock In Kentaro Sano, Dimitrios Soudris, Michael H{\"u}bner, and Pedro~C.
  Diniz, editors, {\em Applied Reconfigurable Computing}, pages 91--102, Cham,
  2015. Springer International Publishing.

\bibitem{7560264}
B.~G. {Sileshi}, J.~{Oliver}, and C.~{Ferrer}.
\newblock Accelerating particle filter on {FPGA}.
\newblock In {\em IEEE Computer Society Annual Symposium on VLSI (ISVLSI)},
  pages 591--594, 2016.
\newblock doi: 10.1109/ISVLSI.2016.66.

\bibitem{vgg16}
Karen Simonyan and Andrew Zisserman.
\newblock Very deep convolutional networks for large-scale image recognition.
\newblock {\em arXiv:1409.1556}, 2014.

\bibitem{resnet}
Kaiming He, Xiangyu Zhang, Shaoqing Ren, and Jian Sun.
\newblock Deep residual learning for image recognition.
\newblock {\em CoRR}, abs/1512.03385, 2015.

\bibitem{iandola2016squeezenet}
Forrest~N Iandola, Song Han, Matthew~W Moskewicz, Khalid Ashraf, William~J
  Dally, and Kurt Keutzer.
\newblock {SqueezeNet}: {AlexNet}-level accuracy with 50x fewer parameters and<
  0.5 mb model size.
\newblock {\em arXiv:1602.07360}, 2016.

\bibitem{iccad}
Yanqi Liu, Alessandro Costantini, Zhiqiang Sui, Zhefan Ye, Shiyang Lu,
  Odest~Chadwicke Jenkins, and R.~Iris Bahar.
\newblock Robust object estimation using generative-discriminative inference
  for secure robotics applications.
\newblock In {\em 2018 IEEE/ACM International Conference on Computer-Aided
  Design (ICCAD)}, pages 1--8, 2018.
\newblock doi:10.1145/3240765.3243493.

\bibitem{importance-sampling}
Malvin~H. Kalos and Paula~A. Whitlock.
\newblock {\em Monte Carlo Methods. Vol. 1: Basics}.
\newblock Wiley-Interscience, USA, 1986.

\bibitem{culling}
Subodh Kumar, Dinesh Manocha, Bill Garrett, and Ming Lin.
\newblock Hierarchical back-face culling.
\newblock In {\em 7th Eurographics Workshop on Rendering}, pages 231--240,
  1996.

\end{thebibliography}
\end{document}